\documentclass{article}
\usepackage{spconf,amsmath,graphicx}
\usepackage{amsthm,amsmath,amssymb}
\usepackage{mathrsfs}
\usepackage{booktabs}

\title{PSFormer: Point Transformer for 3D Salient Object Detection}
\name{Baian Chen$^1$, Lipeng Gu$^1$, Xin Zhuang$^2$, Yiyang Shen$^1$, Weiming Wang$^3$, Mingqiang Wei$^1$}
\address{$^1$Nanjing University of Aeronautics and Astronautics\\ 
$^2$Beijing Aerospace Intelligent Manufacturing Technology Development Co., Ltd\\ 
$^3$Hong Kong Metropolitan University}

\begin{document}
%
\maketitle
\begin{abstract}
We propose PSFormer, an effective point transformer model for 3D salient object detection. 
PSFormer is an encoder-decoder network that takes full advantage of transformers to model the contextual information in both multi-scale point- and scene-wise manners. In the encoder, we develop a Point Context Transformer (PCT) module to capture region contextual features at the point level; PCT contains two different transformers to excavate the relationship among points. In the decoder, we develop a Scene Context Transformer (SCT) module to learn context representations at the scene level; SCT contains both Upsampling-and-Transformer blocks and Multi-context Aggregation units to integrate the global semantic and multi-level features from the encoder into the global scene context. Experiments show clear improvements of PSFormer over its competitors and validate that PSFormer is more robust to challenging cases such as small objects, multiple objects, and objects with complex structures.


\end{abstract}
\section{Introduction}
\label{sec:intro}
While salient object detection (SOD) from 2D images has been extensively studied \cite{borji2015salient,borji2019salient}, there are very few efforts on SOD from 3D point clouds. 
This is despite the fact that the rapid development of 3D acquisition technologies has significantly simplified geometric modeling, and 3D point clouds become more and more popular with wide applications of autonomous driving, and Metaverse. 

\begin{figure}[t]
 \centering
  \includegraphics[width=0.5\textwidth]{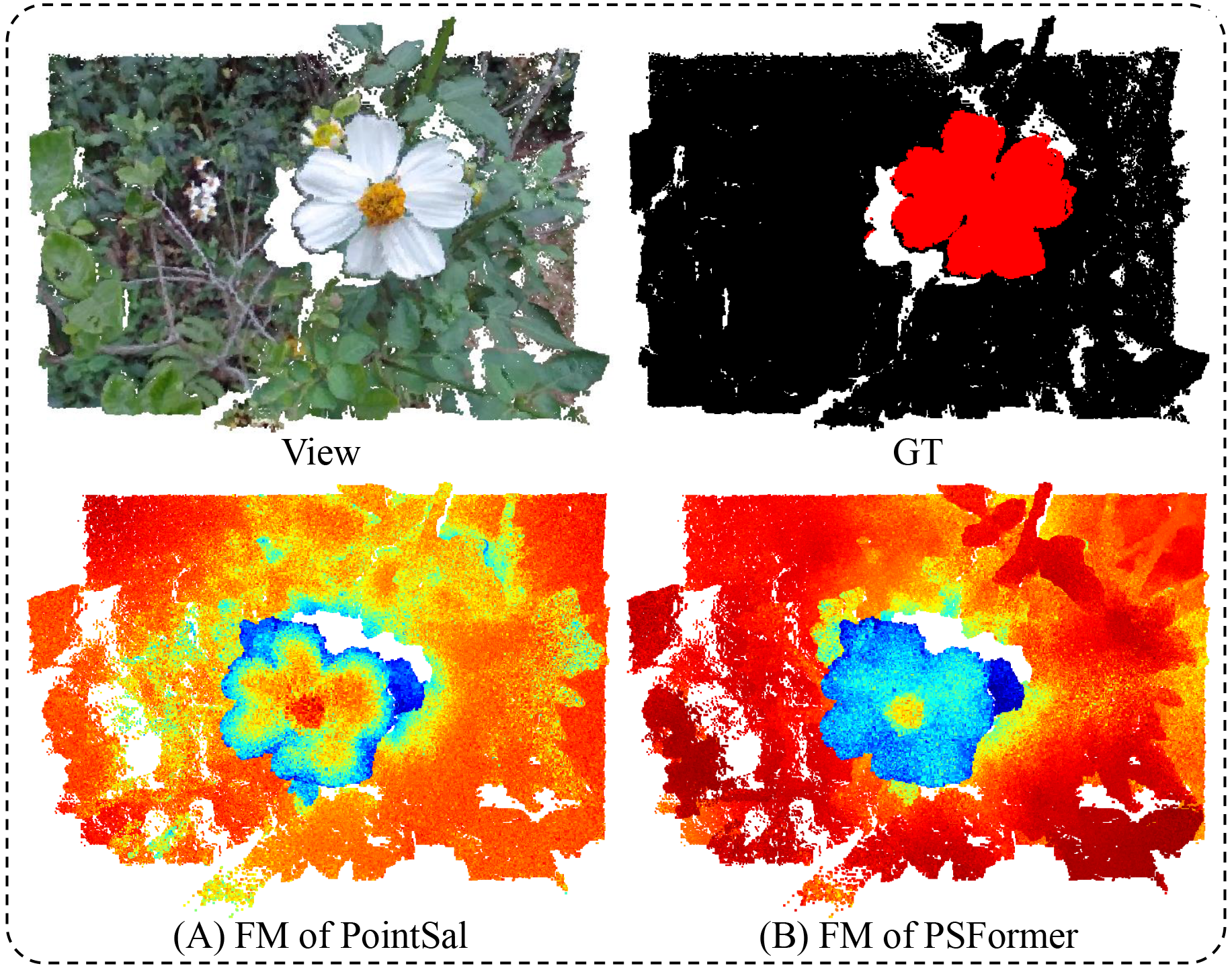}
  \caption{3D heatmap visualization of feature maps (FM). Points that belong to the same semantic part share more similar features by (B) PSFormer than (A) PointSal \cite{fan2022salient}. }
  \label{fig:fig1}
\end{figure}

Unlike images in which salient objects are unchanged, point clouds can be readily rotated. That means, one object may change from the salient to non-salient object during rotation. PointSal \cite{fan2022salient} is the pioneering work of point cloud salient object detection (PCSOD). PointSal takes full advantage of multi-scale features and global semantics to locate salient objects.
However, all feature extraction modules of PointSal are implemented by multi-layer perceptrons (MLPs), which seriously limits the capability of learning long-range feature representations due to fixed receptive fields. When dealing with an object with complex structures (see Figure \ref{fig:fig1} (A)), PointSal \cite{fan2022salient} fails to capture the complete structure information. 


\begin{figure*}[ht]
  \centering
  \includegraphics[width=0.95\textwidth]{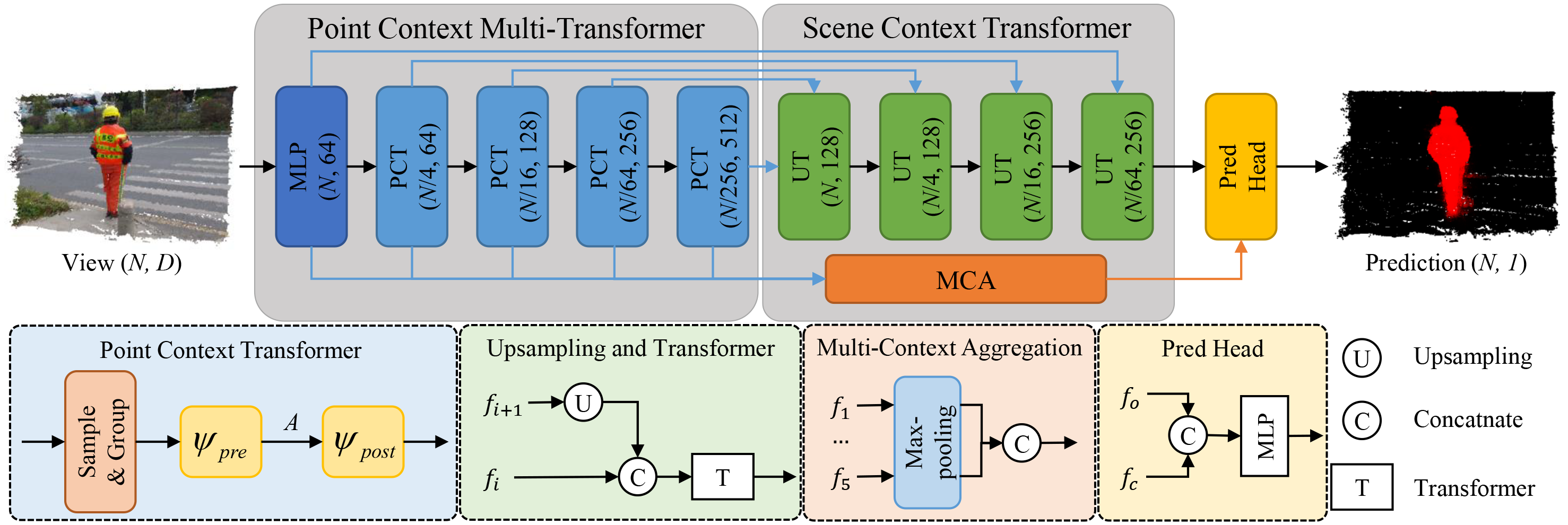}
  \caption{Pipeline of PSFormer. PSFormer is an encoder-decoder structure, mainly consisting of PCT and SCT. 
  }
  \label{fig:fig2}
\end{figure*}

To deal with the challenges of small/multiple objects and objects with complex structures, we propose a point transformer model
for PCSOD, dubbed PSFormer. 
PSFormer leverages multiple transformer modules to capture multi-scale point-wise and scene-wise contextual information simultaneously. In detail,
PSFormer follows an encoder-decoder structure, including two types of transformer modules. The first type is Point Context Transformer (PCT) and the second is Scene Context Transformer (SCT). 
PCT is employed to model the local and global point-wise relationships by two transformers; the local point-wise context can describe the boundary structure and the global point-wise context assists in distinguishing the seed belonging to different classes.
SCT integrates the multi-scale features from decoders to learn the scene-wise context representation; the global scene context can locate the salient object from the interfering background.
Benefiting from PCT and SCT, PSFormer can effectively model the long-range dependencies and learn the contextual information. Therefore, it can be clearly observed in Figure \ref{fig:fig1} (B) that the structural features of the 3D flower are completely distinguished from the complex backgrounds via PSFormer.

To verify the effectiveness of PSFormer, we compare it with PointSal \cite{fan2022salient} and five representative segmentation models on the PCSOD benchmark dataset and achieve the best performance. Our contributions are summarised as follows:\\
(1) We propose a point transformer model
for 3D salient object detection (PSFormer). PSFormer aims to deal with the challenges of small objects, multiple objects, and objects with complex structures. \\
(2) We formulate two types of transformers, i.e., Point Context Transformer and Scene Context Transformer, capturing contextual information at the point and scene levels.\\
(3) Experiments verify that PSFormer outperforms PointSal and five segmentation models on the PCSOD dataset.

\section{Methodology}
\label{sec:method}

\subsection{Overview of PSFormer}
We propose a novel transformer-based salient object detector for point clouds. PSFormer models the context-dependent feature representations in both point and scene levels. It leverages an encoder-decoder structure (see Figure \ref{fig:fig2}). 
The input point cloud is first fed into the encoder to progressively capture high-level semantic features by continuous Point Context Transformer (PCT) blocks. In PCT, two transformers ($\psi_{pre}$ and $\psi_{post}$) learn the point-wise context information in the local and global region. The decoder employs Scene Context Transformer (SCT) module to recover the resolution of the input point cloud by the Upsampling-and-Transformer (UT) block and integrate the scene contextual information via a Multi-Context Aggregation (MCA) module.

\subsection{Point Context Transformer}
To construct the hierarchical feature representation for understanding the semantic information of whole scenes, we follow PointNet++ \cite{NIPS2017_d8bf84be} to build PCT in a pyramid manner. 

Given an input point cloud $\mathcal{P}$ = $\{p_1,p_2,...,p_N\}$, we first generate a new subset $\{p_{c_1},p_{c_2},...,p_{c_M}\}$ by the furthest point sampling (FPS) operation. In this new set, each point is recognized as the centroid to choose $K$ closest points in the local region within a given radius to form the point group.
We denote $\{f_i|i\in{\mathcal{N}(p_{c_t})}\}$ as the $t_{th}$ group with the centroid $p_{c_t}$, where $f_i$
represents features of points in the $t_{th}$ group.
The transformer is formulated as: 
\begin{equation}
\label{eq2}
  q_i=f_iW_q,\quad{} k_i=f_iW_k,\quad{} v_i=f_iW_v,
\end{equation}
\begin{equation}
\label{eq3}
  y_i=Softmax(q_i\cdot{}k_i/\sqrt{d})v_i,
\end{equation}
\begin{equation}
\label{eq4}
  Trans(f_i) = o_i = f_i + FFN(y_i),
\end{equation}
where $W_q$, $W_k$ and $W_v$ are linear projections for query, key and value terms, $d$ represents the feature dimension of key and value vectors, $FFN(\cdot)$ is a feed-forward network.

These groups are first fed into the Feature Normalization (FN) module to normalize the feature distributions in the local region.
Then, the transformer block $\psi_{pre}$ takes these groups as input to model the context dependencies between points in the group, which describes the structural features on the boundary.
The max-pooling function aggregates the features of neighbor points into the centroid to reduce the resolution and expand the receptive field of the sampled seeds.
Subsequently, the aggregated features are fed into the transformer block $\psi_{post}$ to learn the correlations among the sampled seeds, which assists in interfering the category of each point.
The whole operation is formulated as: 
\begin{equation}
  y_i = \psi_{post}(\mathcal{A}(\psi_{pre}(f_{i,j}),|j\in{\mathcal{N}(i)})),
\end{equation}
\begin{equation}
  \psi_{pre}(f_i)=\psi_{post}(f_i)=Trans(f_i).
\end{equation}


\begin{figure}[htbp]
  \centering \includegraphics[width=0.85\linewidth]{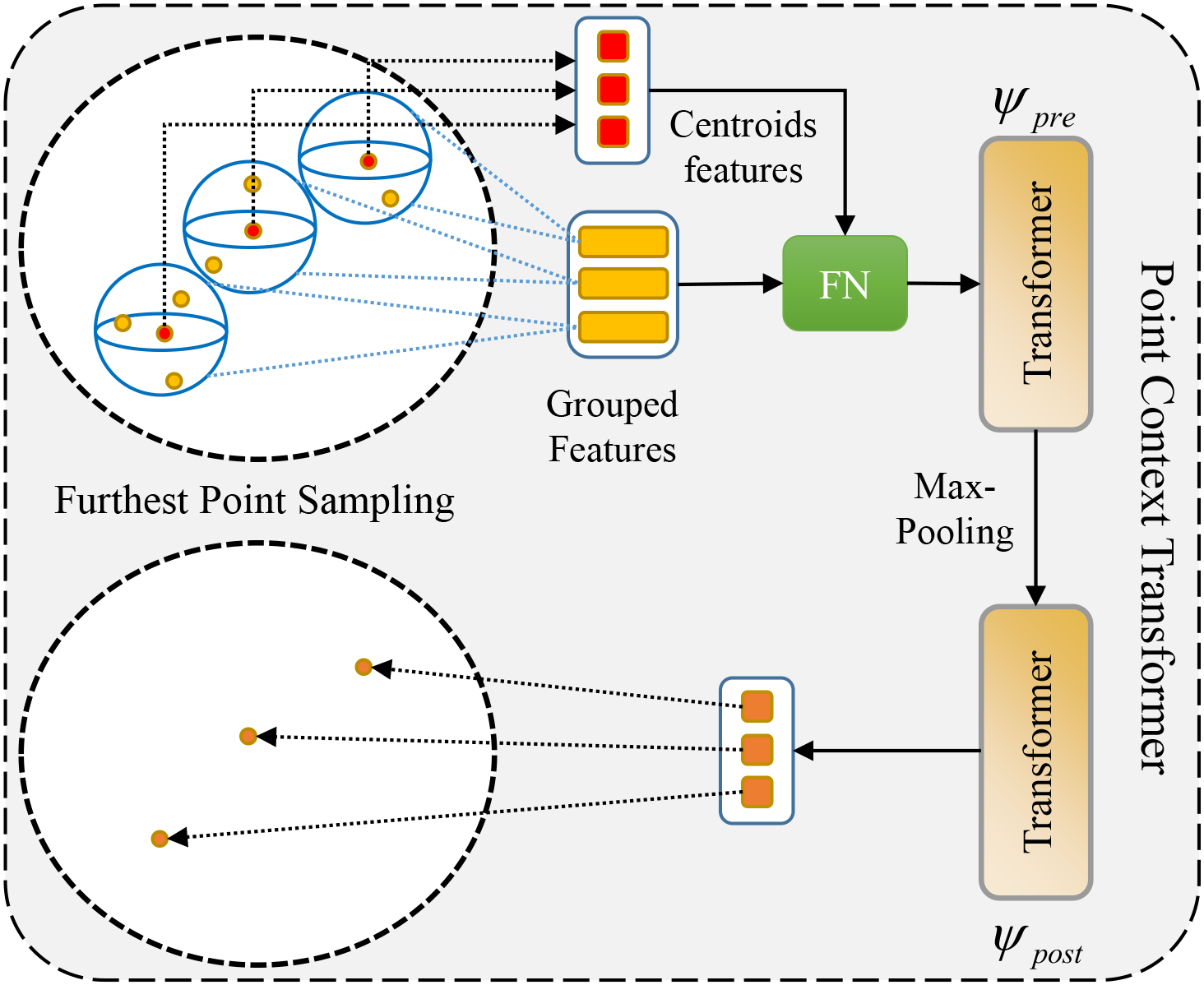}
  \caption{Point Context Transformer (PCT). 
  }
  \label{fig:fig3}
\end{figure}

\begin{figure*}
  \centering
  \includegraphics[width=0.85\textwidth]{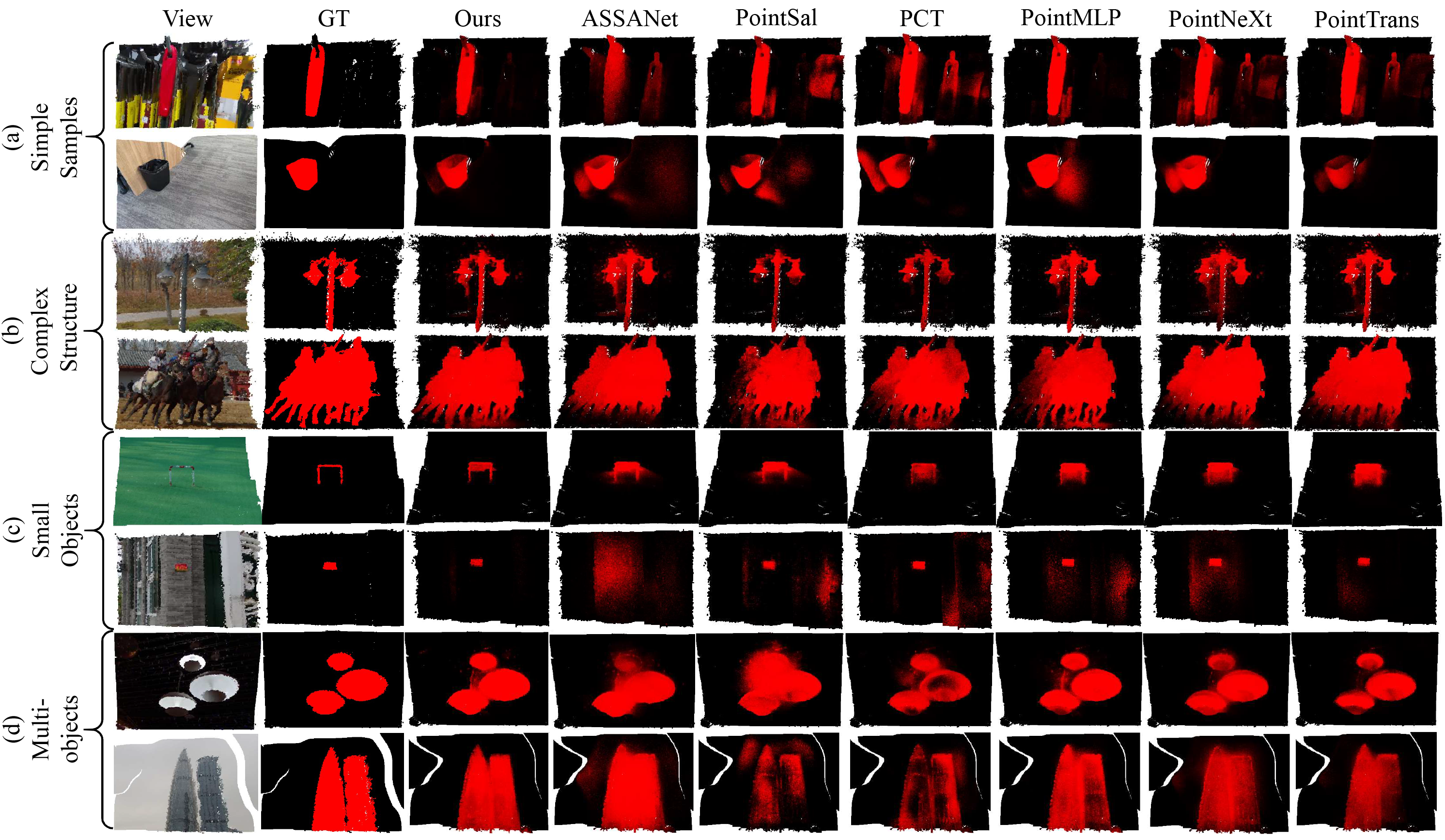}
  \caption{The qualitative of different methods for PCSOD. Our PSFormer behaves better in handling challenging cases, including small objects, multi-objects, and objects with complex structures.}
  \label{fig:fig4}
\end{figure*}

\subsection{Feature Normalization Module}
The self-attention layer is the key component of Transformer, which has a strong ability to learn the feature representations. But simply stacking self-attention layers to learn the deeper features will decrease the accuracy and robustness of performance. This is because points are sparse and irregular and feature distributions among the different local groups are diverse, while shared self-attention layers treat these groups equally.
Inspired by \cite{ma2021rethinking}, we utilize a Feature Normalization (FN) module to assign different weights to diverse groups. Let $\{f_{i,j}\}_{j=1,...,k}\in{\mathbb{R}^{k\times{b}}}$ be the grouped set with the centroid ${f_i}\in{\mathbb{R}^d}$, we transform the local grouped set by 
\begin{equation}
  \sigma = \sqrt{\dfrac{1}{n\times{k\times{d}}}\sum\limits^n_{i=1}\sum\limits^k_{j=1}{(f_{i,j}-f_i)^2}},
\end{equation}
\begin{equation}
  \{f_{i,j}\}=\alpha\odot\dfrac{\{f_{i,j}\}-f_i}{\sigma+\epsilon}+\beta,
\end{equation}
where the standard deviation $\sigma$ of all point cloud features are first calculated to describe the offset across all groups and feature channels, $\alpha$ and $\beta$ are learnable parameters to simulate the distribution in different groups, $\odot$ is dot production and $\epsilon$ is a small number to maintain the numerical stability \cite{ioffe2015batch,wu2018group,dixon1951introduction}.

\begin{table}[t]\footnotesize
\renewcommand\arraystretch{1.0}
  \centering
  \caption{Comparison with the state-of-the-art methods on the PCSOD dataset.}
  \vspace{3pt}
  \resizebox{\linewidth}{!}{
  \begin{tabular}{c|cccc}
  \toprule[1pt]
    Methods & MAE $\downarrow$ & F-measure $\uparrow$ & E-measure $\uparrow$ & IoU $\uparrow$ \\
  \midrule[0.8pt]
    ASSANet \cite{qian2021assanet} & 0.089 & 0.709 & 0.814 & 0.606 \\
    PointTransformer \cite{zhao2021point} & 0.075 & 0.762 & 0.848 & 0.670 \\
    PCT \cite{guo2021pct} & 0.069 & 0.770 & 0.846 & 0.652 \\
    PointMLP \cite{ma2021rethinking} & 0.065 & 0.792 & 0.875 & 0.702 \\
    PointNeXt \cite{qian2022pointnext} & 0.066 & 0.779 & 0.859 & 0.680 \\
  
  \midrule[0.6pt]
    PointSal \cite{fan2022salient} & 0.069 & 0.769 & 0.851 & 0.656 \\
    Ours & \underline{\textbf{0.058}} & \underline{\textbf{0.805}} & \underline{\textbf{0.878}} & \underline{\textbf{0.711}} \\
  \bottomrule[1pt]
  \end{tabular}
  }
  \label{tab:table1}
\end{table}

\subsection{Scene Context Transformer}
The high-level semantics and the multi-scale features are crucial for salient detection tasks \cite{chen2020global,liu2019simple,pang2020multi}.
Thereby we integrate the global semantic and multi-level features $\{F_l\}^5_{l=1}$ from encoders into the global scene context via the Scene Context Transformer (SCT) module. Scene context that describes the object distributions can locate the salient object. 
SCT mainly includes two components, i.e., Upsampling-and-Transformer (UT) and Multi-Context Aggregation (MCA). 

UT upsamples the output $F^{i+1}$ from the previous UT block and concatenates it with features $F^i$ of the corresponding PCT module using the short link. Later, a transformer block excavates the inner relationships of the whole features as 
\begin{equation}
  F_o^i = Trans({C}({U}(F^{i+1}), F^i)),
\end{equation}
where 
$U(\cdot)$ is an upsampling function and $C(\cdot,\cdot)$ is a concatenation function.
MCA directly takes outputs from all encoders as input. We concatenate them together as the global scene context to assist in predicting the salient object. Specifically, we first adopt a channel compression operation for each output, which consists of the MLP layer and Max-pooling. MLP compresses outputs to an identical feature dimension and Max-pooling is employed to generate different vectors for succeeding concatenation as 
\begin{equation}
  F_c = C(Max-pooling(MLP(F^i))),i=1,...,5
\end{equation}

\section{Experiments}
\label{sec:exp}

\subsection{Experimental Setup}
\textbf{Datasets.} PCSOD \cite{fan2022salient} is a benchmark dataset for 3D salient object detection. It includes 2,873 3D views from over one hundred scenes, where each view has 240,000 points. Following the widely used split ratio of 7:3, PCSOD is randomly split into 2,000 training samples and 872 testing samples. Moreover, PCSOD contains a certain amount of challenging samples, such as multiple objects, small objects, complex structures and low illumination. We follow the evaluation protocol proposed along with the dataset.

\textbf{Implementation.} We implement our model with Pytroch on an NVIDIA RTX 2080ti GPU. The point clouds include 9-dimensional features that consist of spatial coordinates, RGB colors and normalized coordinates. We randomly split the complete 3D view into patches with 4,096 points, and treat these patches as input. We train our model by the Adam optimizer in an end-to-end manner. The total training epochs are 800 and the initial learning rate is 5e-4. 

\subsection{Comparison and Analysis}
We compare our method with PointSal \cite{fan2022salient} and five segmentation methods \cite{qian2021assanet,guo2021pct,ma2021rethinking,qian2022pointnext,zhao2021point}. 

\textbf{Quantitative results.} From Table \ref{tab:table1}, our method achieves the best performance among all the methods on PCSOD. 
Our PSFormer outperforms the suboptimal model PointMLP \cite{ma2021rethinking} in all metrics on the PCSOD testing set. 
Point Transformer \cite{zhao2021point} and PCT \cite{guo2021pct} design various transformer modules on the point feature extraction. Although these sophisticated local feature extractors already learn the local context well, their performances are not effective enough due to the lack of representation of the global scene context. PointMLP enhances the ability to learn the point cloud feature representation by stacking more residual feed-forward MLPs. 
But the fixed receptive field restricts the representation power of MLPs.

\textbf{Qualitative results.} To further verify the effectiveness of PSFormer, we visualize the predicted results on some challenging views, $e.g.$, structure-complex objects (Figure \ref{fig:fig4} (b)), small objects (Figure \ref{fig:fig4} (c)) and multi-objects (Figure \ref{fig:fig4} (d)). Our PSFormer can generate more accurate and complete segmentation maps than the other methods.

\begin{table}[t]\footnotesize
\renewcommand\arraystretch{1.1}
  \centering
  \caption{Ablation study on the PCSOD testing set.}
  \vspace{3pt}
  \resizebox{\linewidth}{!}{
    \begin{tabular}{c|cccc}
    \toprule[1pt]
      Methods & MAE $\downarrow$ & F-measure $\uparrow$ & E-measure $\uparrow$ & IoU $\uparrow$ \\
    \midrule[0.8pt]
      PSFormer (w/o FN) & 0.071 & 0.755 & 0.842 & 0.649 \\
      PSFormer (w/o $\psi_{pre}$) & 0.066 & 0.789 & 0.864 & 0.687 \\
      PSFormer (w/o $\psi_{post}$) & 0.063 & 0.799 & 0.873 & 0.700 \\
      PSFormer (w/o UT) & 0.069 & 0.775 & 0.858 & 0.678 \\
      PSFormer (w/o MCA) & 0.061 & 0.801 & 0.873 & 0.702 \\
      
    \midrule[0.6pt]
      PSFormer (full) & \underline{\textbf{0.058}} & \underline{\textbf{0.805}} & \underline{\textbf{0.878}} & \underline{\textbf{0.711}} \\
    \bottomrule[1pt]
  \end{tabular}
  }
  \label{tab:table2}
\end{table}

\subsection{Ablation Study}
We split out five main components from our model, and remove them one by one to verify their effectiveness. They are FN, $\psi_{pre}$, $\psi_{post}$ self-attention layers in PCT and UT, MCA modules in SCT. All results are reported in Table \ref{tab:table2}. It can be observed that the contextual information captured by PCT and SCT indeed improves the performance of our method.

\section{Conclusion}
3D salient object detection is a new topic, remaining many non-trivial problems to solve. For the first time, we propose a transformer model for 3D salient object detection from point clouds, namely PSFormer. Our PSFormer enhances the ability to learn context-dependent feature representations at the point and scene levels by introducing two different-type transformers. The proposed Point
Context Transformer (PCT) models hierarchical context-aware features at the point level by two different transformer blocks. The proposed  Scene Context Transformer (SCT) captures the global scene context by integrating the multi-scale contextual information from different-level PCT modules. Thus, PSFormer is robust to the cases of small objects,
multiple objects, and objects with complex structures. Extensive experiments verify the effectiveness of our method over its competitors.



\bibliographystyle{IEEEbib}
\bibliography{strings,refs}

\end{document}